\documentclass[runningheads]{llncs}
\usepackage{graphicx}
\usepackage{amsmath,amssymb} 
\usepackage{color}
\usepackage{times}
\usepackage{epsfig}
\usepackage{graphicx}
\usepackage{amsmath}
\usepackage{amssymb}
\usepackage{bbm}
\usepackage{marvosym}
\usepackage{subfigure}
\usepackage{algorithm}
\usepackage{algorithmic}
\usepackage{booktabs}
\usepackage{multirow}
\usepackage[table]{xcolor}
\usepackage{colortbl}
\usepackage[pagebackref=true,breaklinks=true,letterpaper=true,colorlinks,bookmarks=false]{hyperref}
\usepackage{breakurl}

\DeclareMathOperator*{\argmax}{arg\,max}

\usepackage[width=122mm,left=12mm,paperwidth=146mm,height=193mm,top=12mm,paperheight=217mm]{geometry}
\begin{document}
\pagestyle{headings}
\mainmatter

\title{``What happens if...''\\Learning to Predict the Effect of Forces in Images}

\titlerunning{}

\authorrunning{Roozbeh Mottaghi, Mohammad Rastegari, Abhinav Gupta, Ali Farhadi}

\author{Roozbeh Mottaghi$^1$, Mohammad Rastegari$^1$, Abhinav Gupta$^{1,2}$, Ali Farhadi$^{1,3}$}
\institute{$^1$Allen Institute for AI, $^2$Carnegie Mellon University, $^3$University of Washington}

\maketitle

\begin{abstract}
\textit{What happens if} one pushes a cup sitting on a table toward the edge of the table? How about pushing a desk against a wall? 
In this paper, we study the problem of understanding the movements of objects as a result of applying external forces to them. For a given force vector applied to a specific location in an image, our goal is to predict long-term sequential movements caused by that force. Doing so entails reasoning about scene geometry, objects, their attributes, and the physical rules that govern the movements of objects. We design a deep neural network model that learns long-term sequential dependencies of object movements while taking into account the geometry and appearance of the scene by combining Convolutional and Recurrent Neural Networks. Training our model requires a large-scale dataset of object movements caused by external forces. To build a dataset of forces in scenes, we reconstructed all images in SUN RGB-D dataset in a physics simulator to estimate the physical movements of objects caused by external forces applied to them. Our Forces in Scenes (ForScene) dataset contains 10,335 images in which a variety of external forces are applied to different types of objects resulting in more than 65,000 object movements represented in 3D. Our experimental evaluations show that the challenging task of predicting long-term movements of objects as their reaction to external forces is possible from a single image. 
\end{abstract}

\section{Introduction}
\begin{figure*}[t]
\centering
  \includegraphics[width=29pc]{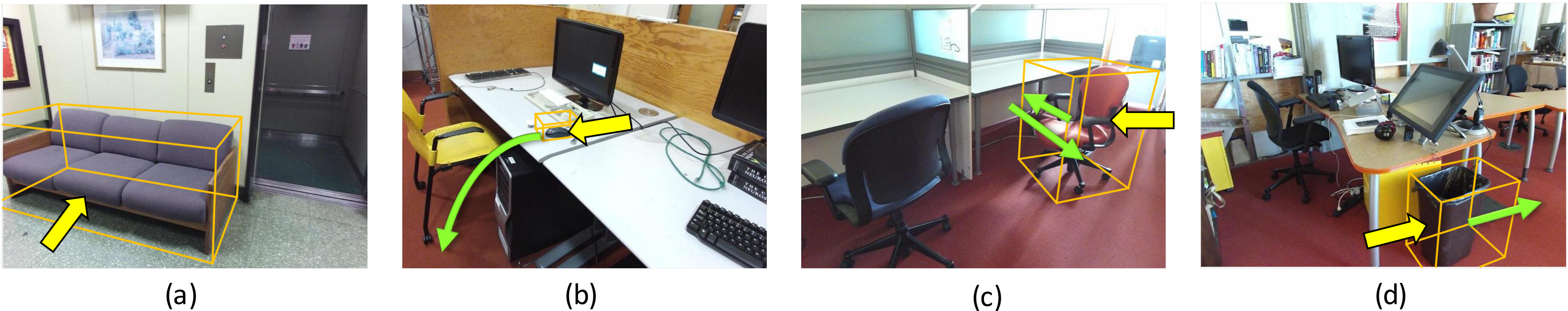}
\caption{\textbf{Subtle differences in forces cause significantly different movements.} The effect of forces on objects depends on the configuration of the scene and object properties. The force is shown in yellow and the direction of movement is shown in green. (a) No movement is caused by the force since there is a wall behind the sofa. (b) The force changes the height of the object. The mouse drops as the result of applying the force. (c) The object might move in the opposite direction of the force. The chair initially moves in the direction of the force, but it bounces back when it hits the desk. (d) The direction of the movement and the force is the same.}
\label{fig:teaser}
\end{figure*}

An important component in visual reasoning is the ability to understand the interaction between forces and objects; and the ability to predict the movements caused by those forces. We humans have an amazing understanding of how  applied and action-reaction forces work. In fact, even with a static image \cite{hamrick11,battaglia13}, humans can perform a mental simulation of the future states and reliably predict the dynamics of the interactions. For example, a person can easily predict that the couch in Figure~\ref{fig:teaser}(a) will not move if it is pushed against the wall and the mouse in Figure~\ref{fig:teaser}(b) will eventually drop if it is pushed towards the edge of a desk. 

\begin{figure*}[t]
\centering
  \includegraphics[width=29pc]{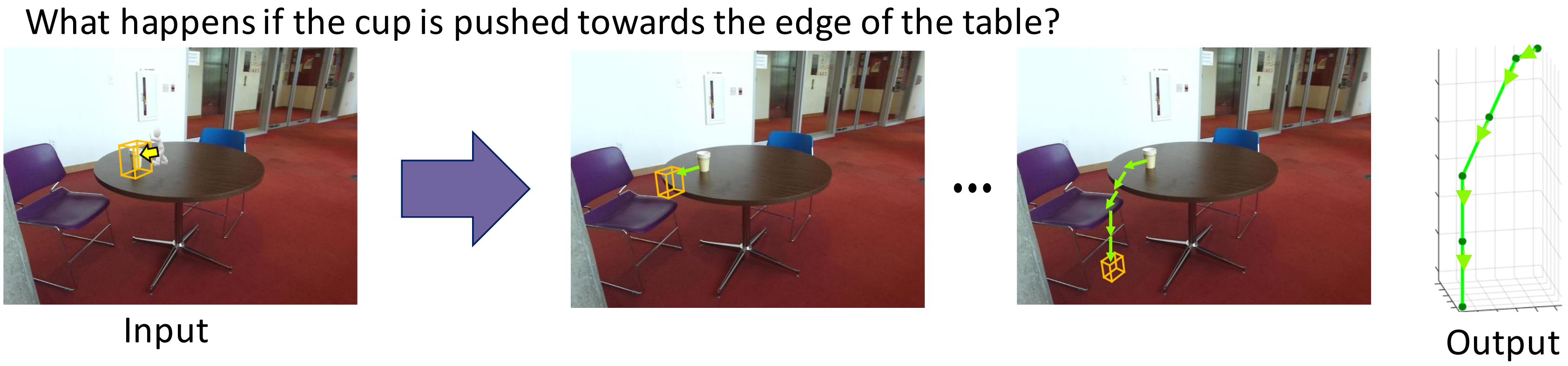}
\caption{Our goal is to learn ``What happens if Force X is applied to Point Y in the scene?". For example, from a single image, we can infer that the cup will drop if we push it towards the edge of the table. On the right we show the output of our method, i.e. a sequence of velocity vectors in 3D which are caused by applying the force.}
\label{fig:pullfig}
\end{figure*}

In this paper, we address the problem of predicting the effects of external forces applied to an object in an image. Figure~\ref{fig:pullfig} shows a long-term prediction of the sequence of movements of a cup when it is pushed toward the edge of the table. Solving this problem requires reliable estimates of the scene geometry, the underlying physics, and the semantic and geometric properties of objects. Additionally, it requires reasoning about interactions between forces and objects where subtle changes in how the force is applied might cause significant differences in how objects move. For example, depending on the magnitude of the force, the cup remains on the table or falls. What makes this problem more challenging is the sequential nature of the output where predictions about movements of objects depend on the estimates from the previous time steps. Finally, a data-driven approach to this problem requires a large-scale training dataset that includes movements of objects as their reaction to external forces. Active interaction with different types of scenes and objects to obtain such data is non-trivial.

Most visual scene understanding methods (e.g., \cite{murphy03,gupta10,yao12}) are \emph{passive} in that they are focused on predicting the scene structure, the objects, and their attributes and relations. These methods cannot estimate what happens if some parts of the scene are changed actively. For example, they can predict the location or 3D pose of a sofa, but they cannot predict how the sofa will move if it is pushed from behind. In this paper, we focus on an \emph{active} setting, where the goal is to predict \textit{``What happens if Force X is applied to Point Y in the scene?''} 

We design a deep neural network model that learns long-term sequential dependencies of object movements while taking into account the geometry and appearance of the scene by combining Convolutional and Recurrent Neural Networks. The RNN learns the underlying physical rules of movements while the CNN implicitly encodes the appearance and geometry of the object and the scene.
To obtain a large number of observations of forces and objects to train this model, we collect a new dataset using physics engines; current datasets in the literature represent static scenes and are not suitable for active settings. Instead of training our model on synthetic images we do the inverse: we replicate all the scenes of SUN RGB-D dataset \cite{sunrgbd} in a physics engine. The physics engine can then simulate forward the effect of applying forces to different objects in each image. We use the original RGB images, the forces, and their associated movements to form our dataset for training and evaluation.

Our experimental evaluations show that the challenging task of predicting long-term movements of objects as their reaction to external forces is possible from a single image.  Our model obtains promising results in predicting the direction of the velocity of objects in 3D as the result of applying forces to them. We provide results for different variations of our method and show that our model outperforms baseline methods that perform regression and nearest neighbor search using CNN features. Furthermore, we show that our method generalizes to object categories that it has not seen during training.

\section{Related Work}
\textbf{Passive scene understanding.} There is a considerable body of work on scene understanding in the vision literature, e.g., \cite{murphy03,gupta10,yao12,li09,heitz08,silberman12,choi13,lin13,zhang14}. However, most of these works propose \emph{passive} approaches, where they infer the \emph{current} configuration of the scenes (location of the objects, their 3D pose, support relationship, etc.) depicted in images or videos. In contrast, our method is an \emph{active} approach, where we predict the result of interacting with objects using forces. 

\textbf{Physics-based prediction.} \cite{mottaghi16} infer the dynamics of objects from a single image. They infer the force using a data-driven approach. In this work, we explicitly represent the force in the scene and predict the object movements. \cite{Fragkiadaki16} predict the effect of forces in a billiard scene. Our method infers the movements based on a single image, while \cite{Fragkiadaki16} uses a sequence of images. Also, \cite{Fragkiadaki16} works on synthetic billiard scenes, while our method works on realistic images. \cite{zheng14} detect potentially falling objects given a point cloud representing the scene. In contrast, our method is based solely on visual cues and does not explicitly use physics equations.

\textbf{Estimating physical properties.} \cite{Bhat2002} estimate the physical parameters of rigid objects using video data. \cite{brubaker2009} estimates forces applied to a human using the dynamics of contacts with different surfaces. \cite{wu15} learn a model for estimating physical properties of objects such as mass and friction based on a series of videos that show movement of objects on an inclined surface. These methods are not designed to predict the result of applying new forces to the scene and are limited to their controlled settings. 

\textbf{Stability inference.} \cite{zheng13} reasons about the stability of objects in a given point cloud. \cite{jia13} solves a joint optimization for segmentation, support relationships and stability. \cite{jiang12} propose a method to place a new object in a stable and semantically preferred location in a scene. Our method, in contrast, predicts the future movements of objects caused by applying forces to them.

\textbf{Predicting sequences using neural networks.} \cite{ranzato14} propose a recurrent architecture to predict future frames of a video. \cite{oh15} propose a recurrent neural net to predict the next frame in an Atari game given the current action and the previous frames. \cite{sutskever08} propose Recurrent RBMs to model high dimensional sequences. \cite{michalski14} model temporal dependencies of a sequence and predict multiple steps in the future. These approaches either require a full sequence (past states and current actions) or work only on synthetic data and in limited environments. Also, \cite{levine15} propose a deep-learning based method to perform a pre-defined set of tasks. They learn a distribution over actions given the current observation and configurations. In contrast, we predict how the scene changes as the result of an action (i.e. applying forces). In the language domain, \cite{karpathy15,vinyals15} have used a combination of CNNs and RNNs to generate captions for images.

\textbf{Data-driven prediction.} \cite{walker14} infers the future path of rigid objects according to learned models of appearance, context, and transition. \cite{pintea14,walker15} predict optical flow from a single image. \cite{yuen10} predict future events that might take place in a query image. \cite{kitani12} estimate future movements of humans in a given scene. \cite{fouhey14} predicts relative movements of objects. Unlike these approaches, we explicitly represent forces and focus on the physics of the scene in order to infer future movements of objects in 3D.

\textbf{Physics-based tracking.} \cite{salzmann11} recover 3D trajectories and the forces applied to objects in a tracking framework. \cite{vondrak08} incorporates physical plausibility into a human tracking framework. Our problem is different from tracking since we perform inference using a single image.

\section{Problem Statement}
\label{sec:problem}
Given a query object in a single RGB image and a force vector, our goal is to predict the future movement of the object as the result of applying the force to the object. More specifically, for a force $\vec{f}$ and an impact point $\vec{p}$ on the object surface in the RGB image, our goal is to estimate a variable-length sequence of velocity directions $V=(v_0, v_1, \ldots, v_t)$ for the center of the mass of the object. These velocities specify how the location of the object changes over time.  

For training we need to obtain the sequence $V$ that is associated to force $\vec{f}=(f_x,f_y,f_z)$ applied to point $\vec{p}=(p_u,p_v)$ in the image. To this end, we automatically synthesize the scene in a physics engine (described in Section~\ref{sec:dataset}). The physics engine simulates forward the effect of applying the force to the point that corresponds to $\vec{p}$ in the 3D synthetic scene and generates the velocity profile and locations for the query object.

During testing, we do not have access to the synthesized scene or the physics engine, and our goal is to predict the sequence $V$ given a query object in a single RGB image and a force\footnote{We refer to the force and its impact point as \emph{force} throughout the paper.}.

We formulate the estimation of the movements as a sequential classification problem. Hence, each $v_t$ takes a value from the set $\mathcal{L}=\{l_1, l_2,\ldots, l_N, s \}$, where each $l_i$ denotes the index for a direction in the quantized space of 3D directions, and $s$ represents `stop' (no motion). The velocity at each time step $v_t$ depends on the previous movements of the object. Therefore, a natural choice for modeling these temporal dependencies is a recurrent architecture. To couple the movement information with the appearance and geometry of the scene and also the force representation, our model integrates a Recurrent Neural Network (RNN) with a Convolutional Neural Network (CNN). Section~\ref{sec:model} describes the details of the architecture.

\begin{figure*}[t]
\centering
  \includegraphics[width=29pc]{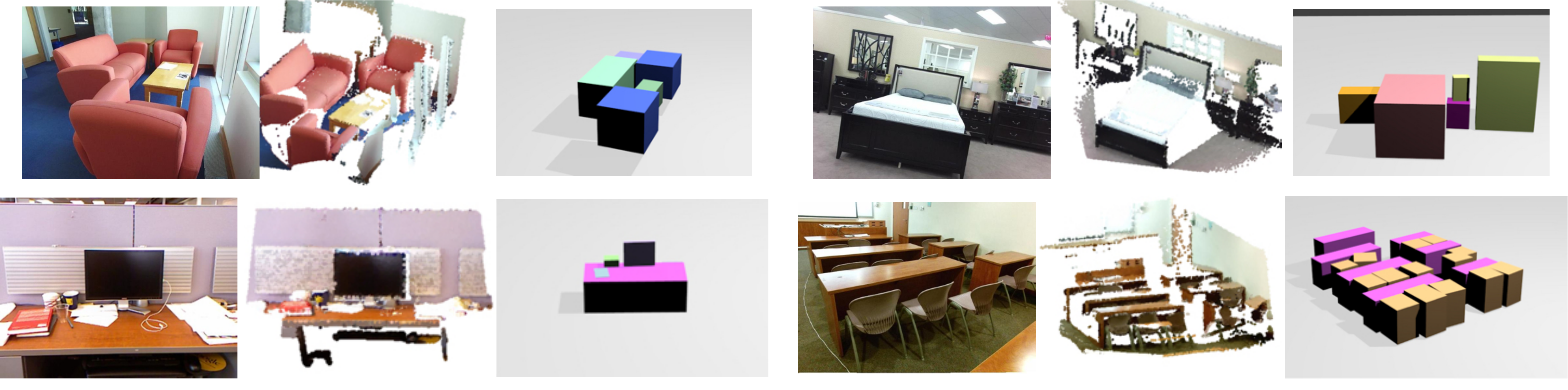}
\caption{\textbf{Synthetic scenes.} These example scenes are synthesized automatically from the images in the SUN RGB-D \cite{sunrgbd} dataset. Left: the original image, Middle: point cloud representation, Right: synthetic scene. The objects that belong to the same category are shown with the same color. For clarity, we do not visualize the walls.}  
\label{fig:dataset}
\end{figure*} 

\section{Forces in Scenes (ForScene) Dataset} 
\label{sec:dataset}
One of the key requirements of our approach is an \emph{interactable} dataset. Most of the current datasets in the vision community are `static' datasets in that we cannot apply forces to objects depicted in the scenes and modify the scenes. For example, we cannot move the constituent objects of the scenes shown in PASCAL \cite{everingham10} or COCO \cite{lin14} images as we desire since inferring the depth map and the physics of the scene from a single RGB image is a challenging problem. An alternative would be to use RGB-D images, where the depth information is available. This solves the problem of depth estimation and moving the objects in perspective, but RGB-D images do not provide any information about the physics of the world either. 

To make an \emph{interactable} dataset, we transfer the objects and the scene layout shown in images to a physics engine. The physics engine takes a scene and a force as input and simulates the future states of the objects in the scene according to the applied forces. This enables us to collect the velocity sequences that we require for training our model.

Our dataset is based on the SUN RGB-D dataset \cite{sunrgbd}. The SUN RGB-D dataset includes dense 2D and 3D annotations for 10,335 images. These annotations are in the form of 3D bounding boxes and 2D semantic segmentation for about 1,000 object categories. The 3D position and orientation of each bounding box is provided in the annotations, hence, we can transfer the 3D bounding boxes of the objects to the physics engine and reconstruct the same object arrangement in the physics engine. In addition, the SUN RGB-D dataset includes annotations for the scene layout (floors, walls, etc). We replicate the scene layout in the physics engine as well. Figure~\ref{fig:dataset} shows a few examples of the images and their corresponding scenes in the physics engine. We could alternatively use other scene datasets to construct our physics engine scenes, but those datasets were either small \cite{silberman12} or non-photo-realistic \cite{scenenet}. More details about the dataset can be found in Section~\ref{sec:subdataset}. Note that our training and evaluation is performed on real images. These synthetic scenes only supply the groundtruth velocity information.

\section{Model}
\label{sec:model}
We now describe different components of our model, how we represent objects and forces in the model and how we formulate the problem to predict the movements. 
\begin{figure*}[t]
\centering
  \includegraphics[width=29pc]{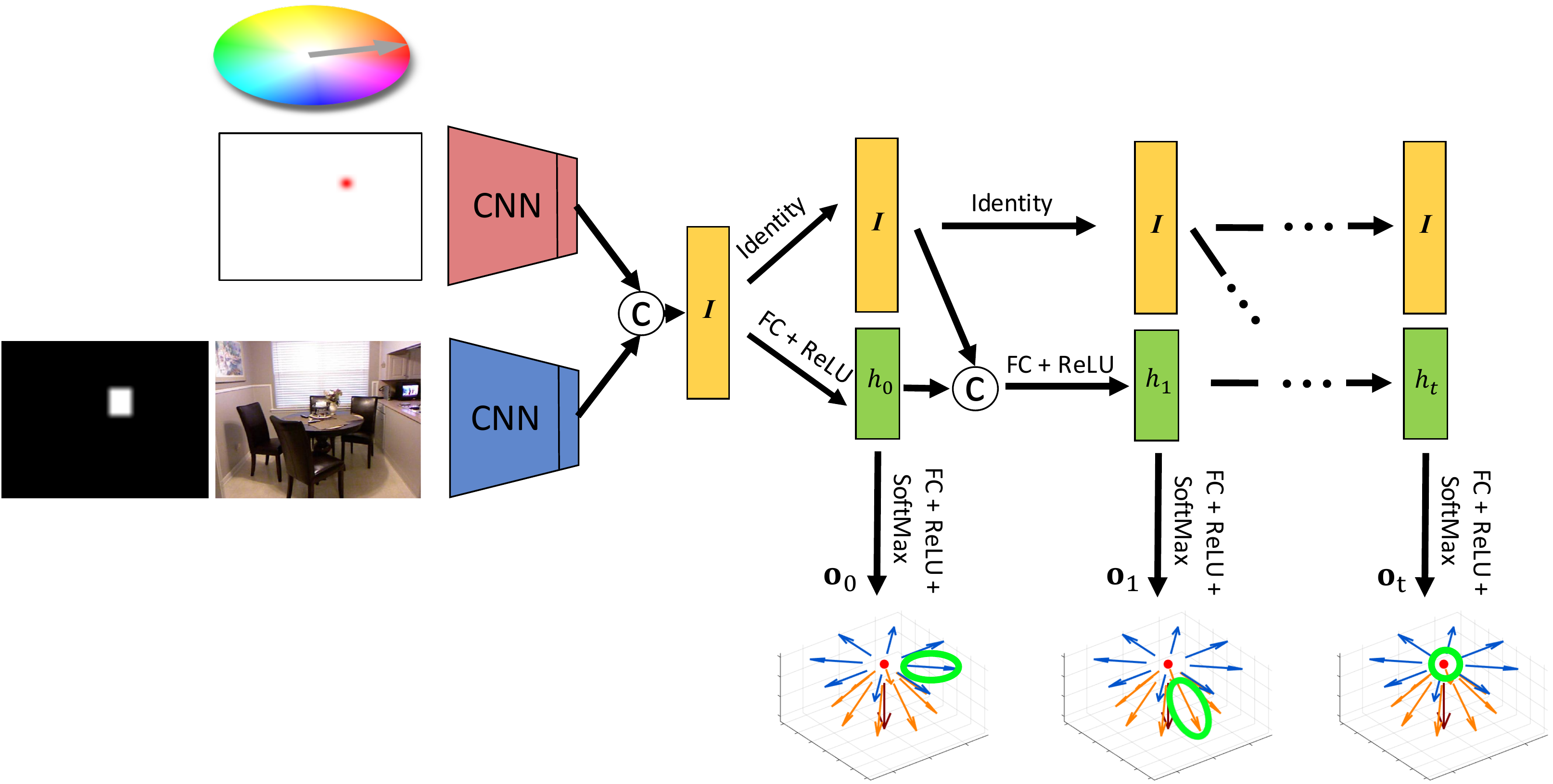}
\caption{\textbf{Model.} Our model consists of two CNNs for capturing the \emph{force} and \emph{image} information. We refer to these CNNs as force tower and image tower respectively. The input to the model is a force image and an RGB-M image (RGB image plus an M channel representing object bounding box). The color in the force image represents the direction and magnitude of the force (according to the color wheel). The symbol $\copyright$ denotes concatenation. `Identity' propagates the input to the output with no change. $h_t$ represents the hidden layer of the RNN at time step $t$. Also, we use the abbreviation FC for a fully connected layer. The output of our model is a sequence of velocity \emph{directions} at each time step. We consider 17 directions and an additional `stop' class, which is shown by a red circle. The green ellipses show the chosen direction at each time step. The RNN stops when it generates the `stop' class.}
\label{fig:model}
\end{figure*}

\subsection{Model architecture}
Our model has three main components: (1) A Convolutional Neural Network (CNN) to encode scene and object appearance and geometry. We refer to this part of the model as \emph{image tower}. (2) Another CNN (parallel to the image tower) to capture force information. We refer to this part of the model as \emph{force tower}. (3) A Recurrent Neural Network (RNN) that receives the output of the two CNNs and generates the object motion (or equivalently, a sequence of vectors that represent the velocity of the object at each time step). Note that the training is end-to-end and is performed jointly for all three components of the model. Figure~\ref{fig:model} illustrates the architecture of the full network. 

We use two different architectures for the image tower for the experiments: AlexNet \cite{alexnet} and ResNet-18 \cite{he16}, where we remove their final classification layer. Similar to  \cite{mottaghi16}, the input to our image tower is a four-channel RGB-M image, where we add a mask channel (M) to the RGB image. The mask channel represents the location of the query object and it is obtained by applying a Gaussian kernel to a binary image that shows the bounding box of the query object. We propagate the output of the layer before the last layer of the CNN (e.g., FC7 when we use AlexNet) to the next stages of the network.

The force tower is structured as an AlexNet \cite{alexnet} and is parallel to the image tower. The input to the force tower is an RGB image that represents the impact point, direction and magnitude of the query force (we will explain in Section~\ref{sec:forcerep} how this image is created). The output of the FC7 layer of the force tower is propagated to the next stages of the network. Our experiments showed that using a separate force tower provides better results compared to adding the force as another input channel to the image tower. Probably, the reason is that there is too much variability in the real images, and the network is not able to capture the information in the force image when we have a single tower for both real images and force images. Therefore, we consider two separate towers and combine the output of these towers at a later stage. The outputs of the image tower and force tower are concatenated (referred to as $I$ in Figure~\ref{fig:model}) and provide a compact encoding of the visual cues and force representation for the recurrent part of the network. 

The recurrent part of our network receives $I$ as input and generates a sequence of velocity vectors. The advantage of using a Recurrent Neural Network (RNN) is two-fold. First, the velocities at different time steps are dependent on each other, and the RNN can capture these temporal dependencies. Second, RNNs enable us to predict a variable-length sequence of velocities (the objects move different distances depending on the magnitude of the force and the structure of the scene). We show the unfolded RNN in Figure~\ref{fig:model}. The hidden layer of the RNN at time step $t$ is a function of $I$ and the previous hidden unit ($h_{t-1}$). More formally, $h_t = f(I,h_{t-1})$, where $f$ is a linear function (fully connected layer) followed by a non-linear ReLU (Rectified Linear Unit). \cite{le15} show that RNNs composed of ReLUs and initialized with identity weight matrix are as powerful as standard LSTMs. The first hidden unit of the RNN ($h_0$) is only a function of $I$. The output at each time step $\mathbf{o}_t$ is a function of the hidden layer $h_t$. More concretely, $\mathbf{o}_t = \mbox{SoftMax}(g(h_t))$, where $g$ is a linear function, which is augmented by a ReLU. 

We use 1000 neurons for the hidden layer in the recurrent part of the network. The output $\mathbf{o}_t$ is of size $|\mathcal{L}|$. $\mathcal{L}$, as defined in Section~\ref{sec:problem}, is a set of directions in 3D and a `stop' class, which represents the end of the sequence. Note that the input to the RNN, $I$, remains the same across different steps of the RNN. 

\subsection{Training}
To train our model, in each iteration, we feed a random batch of RGB-M images from the training set into the image tower. The corresponding batch of force images is fed into the force tower. There is a sequence of velocity vectors associated to each pair of RGB-M and force images. These sequences have different lengths depending on the velocity profile of the query object in the groundtruth. If the object does not move as the result of applying the force, the sequence will be of length 1, where its value is `stop'. The training is performed end-to-end, and each iteration involves a forward and a backward pass through the entire network. 

The loss function is defined over the sequence of outputs $O=(\mathbf{o}_0, \mathbf{o}_1,\ldots,\mathbf{o}_{t^\prime})$. Suppose the groundtruth velocity sequence is denoted by $V=(v_0, v_1, \ldots, v_t)$, the classification loss, $E(V, O)$, which is based on the cross entropy loss, is defined as follows:
\begin{equation}
     E(V, O) = -\frac{1}{T}\sum_{t = 0}^T q_t(v_t)\,\, \mbox{log}(\mathbf{o}_t[v_t]),
\end{equation}
where $\mathbf{o}_t[v_t]$ represents the $v_t$-th element of $\mathbf{o}_t$, $T$ is the maximum length of a sequence, and $q_t(v_t)$ is the inverse frequency of direction $v_t$ in step $t$ of the sequences in the training data. We pad the end of the sequences whose length is shorter than $T$ (i.e. $|O| < T$ or $|V| < T$) with `stop' so their length becomes equal to $T$. We could alternatively represent velocities as 3-dimensional vectors and use a regression loss instead. However, we achieved better performance using the classification formulation. A similar observation has been made by \cite{wang15,walker15} that formulate a continuous variable estimation problem as classification.

\subsection{Testing}
The procedure for predicting a sequence of velocity vectors is as follows. We obtain $I$ (the input to the RNN) by feeding the RGB-M and force images into the object and force towers, respectively. The hidden unit $h_0$ is computed according to the fully connected layer that is defined over $I$. The first velocity in the sequence, $v_0$, is computed by taking the argmax of the output of the SoftMax layer that is defined over $h_0$. We compute $h_1$ based on $I$ and $h_0$ and similarly find the next velocity, $v_1$, in the sequence. More concretely, $v_t = \argmax \mathbf{o}_t$ (recall that $v_t$ is the index for a direction in the quantized set of directions or `stop'). We continue this process until the RNN generates the `stop' class (i.e. $v_t=\mbox{stop}$) or it reaches the maximum number of steps that we consider. 

\section{Experiments}

In this section, we describe the evaluation of our method and compare our method with a set of baseline approaches. We provide the details of the dataset and explain how we interact with objects in the scenes. Additionally, we explain how we represent the force in the CNN and provide more implementation details about our network. To ensure the reproducibility of these experiments, we plan to release the code and the dataset. 

\subsection{Dataset details}
\label{sec:subdataset}
Our dataset is based on the SUN RGB-D \cite{sunrgbd} dataset, which contains 10,335 images (divided into 2666, 2619 and 5050 images for training, validation, and test, respectively). Each object is annotated with a 3D bounding box (position and orientation in 3D) and a segmentation mask (a 2D segmentation mask in the RGB image). There are more than 1,000 object categories in the dataset. Additionally, the room layout annotations (in the form of walls and floors) are provided for each image in the dataset. These annotations enable us to automatically reconstruct a similar scene in a physics engine. Therefore, for each image in \cite{sunrgbd}, we have a synthetic scene, which will be used to simulate the effect of the forces.

We use Blender physics engine\footnote{http://www.blender.org} to render the synthetic scenes. Some example scenes and their corresponding images are shown in Figure~\ref{fig:dataset}\footnote{Some objects are missing in the synthetic scenes since they are not annotated in the SUN RGB-D dataset.}. To create our dataset, we use all $\sim$1,000 categories and walls and floors to construct the synthetic scene, however, we apply the force to the 50 most frequent rigid categories in the dataset. These categories include: chair, keyboard, flower vase, etc. The full list of the 50 categories is in Appendix. We represent each object as a cube in the synthetic scene. 

For each object in the image, we randomly select a point on the surface of the object and apply the force to this point (note that during training for each point in the RGB image, we know the corresponding 3D point in the synthetic scene). The input force is also chosen at random. We simulate the scene after applying the force to the impact point. The simulation continues until the object to which the force is applied reaches a stable state, i.e. the linear and angular velocities of the object become zero. Over the entire dataset, it took a maximum of 32 simulation steps that the object converges to the stable position. We sample velocities every 6 steps, which results in a sequence of at most 6 velocity vectors (depending on the number of steps needed for convergence to stability). We use this sequence as the groundtruth sequence for the query object and force. We represent these velocities in a quantized space of 3D directions (we ignore the magnitude of the velocities), where the directions are 45 degrees apart from each other. Figure~\ref{fig:model} shows these directions. We have 17 directions in total, hence, the size of the set $\mathcal{L}$ (defined in Section~\ref{sec:problem}) will be 18 (17 directions + 1 `stop' class). We assign the velocity vector to the nearest direction class using angular distance. If the magnitude of the velocity vector is lower than a threshold we assign it to the `stop' class. These directions cover a semi-sphere since the velocity directions in the other semi-sphere are rare in our dataset.

\begin{figure*}[t]
\centering
  \includegraphics[width=29pc]{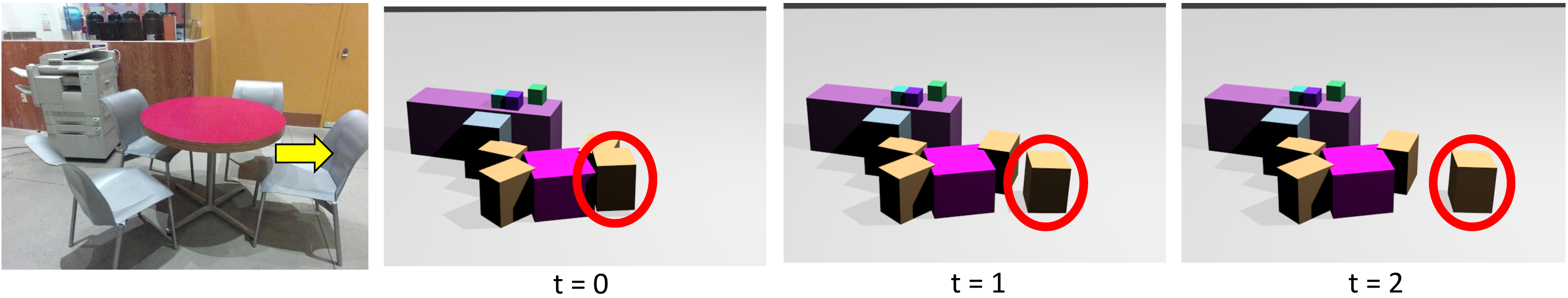}
\caption{\textbf{Synthesizing the effect of the force.} A force (shown by a yellow arrow) is applied to a point on the surface of the chair. The three pictures on the right show different time steps of the scene simulated in the physics engine. There is a red circle around the object that moves.}
\label{fig:gevideo}
\end{figure*}

As the result of the simulations, we obtain 30,655 velocity sequences for training and validation and 34,777 sequences for test. Note that sometimes we apply the force in the same direction but with different magnitudes. In the real world, some of the objects such as toilets or kitchen cabinets are fixed to the floor. We consider those object categories as `static' in the physics engine, which means we cannot move them by applying a force. Figure~\ref{fig:gevideo} shows an example sequence of movements in a synthetic scene. 

\subsection{Force representation}
\label{sec:forcerep}
To feed the force to the CNN, we convert the force vector to an RGB image. Here we describe the procedure for creating the force image. For simplicity, when we collect the dataset, we set the z component of our forces to zero (we refer to the axis that is perpendicular to the ground as the z axis). However, note that the z component of their corresponding velocities can be non-zero (e.g., a falling motion). The force image is the same size as the input RGB image. We represent the force as a Gaussian that is centered at the impact point of the force in the 2D image. We use a color from a color wheel (shown in Figure~\ref{fig:model}) to represent the direction and the magnitude of the force. Each point on the color wheel specifies a unique direction and magnitude. The standard deviation of the Gaussian is 5 pixels in both directions.

\subsection{Network and optimization parameters}

We used Torch\footnote{http://torch.ch} to implement the proposed neural network. We run the experiments on a Tesla K40 GPU. We feed the training images to the network in batches of size 128 when we use AlexNet for the image tower and of size 96 when we use ResNet-18 for the image tower. Our learning rate starts from $10^{-2}$ and gradually decreases to $10^{-4}$. We initialize the image tower and the force tower by a publicly available AlexNet model\footnote{http://github.com/BVLC/caffe/tree/master/models/bvlc\_alexnet} or ResNet model\footnote{https://github.com/facebook/fb.resnet.torch/tree/master/pretrained} that are pre-trained on ImageNet. We randomly initialize the 4th channel of the RGB-M image (the M channel) by a Gaussian distribution with mean 0 and standard deviation 0.01. The forward pass and the backward pass are performed for 15,000 iterations when we use AlexNet for the image tower (the loss value does not change after 15K iterations). When we use ResNet-18 we use 35,000 iterations since it takes longer to converge.

\subsection{Prediction of velocity sequences}
We evaluate the performance of our method on predicting the 34,777 sequences of velocity vectors in the test portion of the dataset.

\noindent \textbf{Evaluation criteria.}
To evaluate the performance of our method, we compare the estimated sequence of directions with the groundtruth sequence. If the predicted sequence has a different length compared to the groundtruth sequence, we consider it as incorrect. If both sequences have the same length, but they differ in at least one step, we consider that as an incorrect prediction as well. We report the percentage of sequences that we have predicted entirely correctly. We have about 1000 patterns of sequences in our test data so the chance performance is close to 0.001.

\begin{figure}[t]
\centering
\subfigure[]{%
\includegraphics[width=0.45\textwidth]{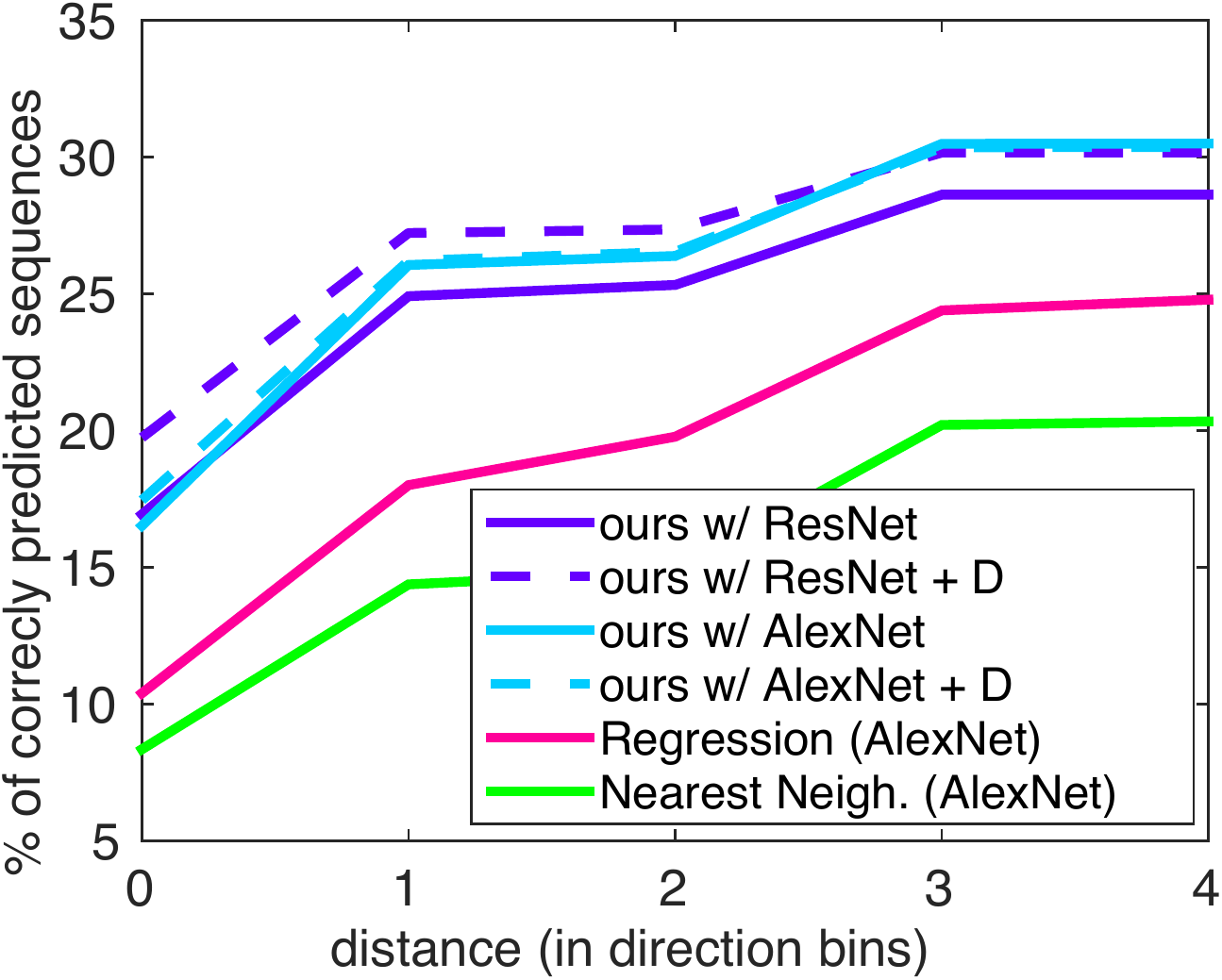}
\label{fig:knear}}
\subfigure[]{%
\includegraphics[width=0.47\textwidth]{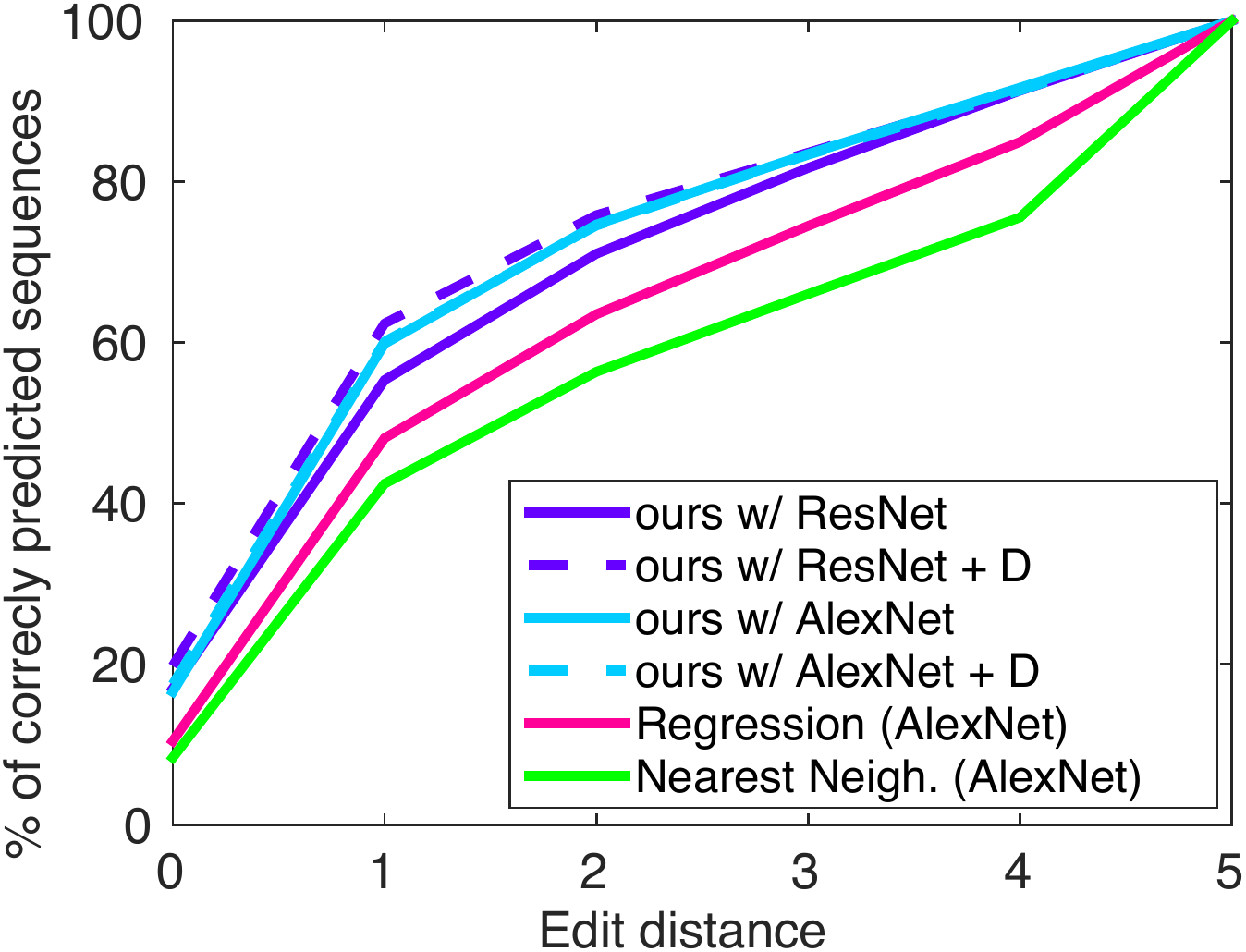}
\label{fig:editdist}}
\caption{\textbf{Relaxation of the evaluation criteria.} (a) We consider the prediction for each step as correct if it is among the $k$ nearest directions to the groundtruth direction. The x-axis shows $k$. (b) We consider a predicted sequence as correct if it is within edit distance $k$ of the groundtruth sequence. The x-axis shows $k$.}
\label{fig:result}
\end{figure}

\noindent \textbf{Results.} We estimate 16.5\% of the sequences in the test data  correctly using our method that uses AlexNet as image and force towers. We refer to this method as `ours w/ AlexNet' in Table~\ref{tab:results}. The criteria that we consider is a very strict criteria. Therefore, we also report our results using less strict criteria. We consider a direction as correct if it is among the closest $k$ directions to the groundtruth direction. Figure~\ref{fig:knear} shows these results for $k=0,\ldots,4$ ($k=0$ means we compare with the actual groundtruth class). We observe a significant improvement using this relaxed criteria. We also report the results using `edit distance', which is a measure of dissimilarity between the groundtruth and the predicted sequences. Basically, it measures how many operations we need to convert a sequence to the other sequence. We report what percentage of predicted sequences are correct within edit distances 0 to 5. This result is shown in Figure~\ref{fig:editdist}. The result of `ours w/ AlexNet' improves to 59.8\% from 16.5\% if we consider the predictions whose edit distance with the groundtruth is less than or equal to 1, as correct.

We also replaced the AlexNet in the image tower by the ResNet-18 \cite{he16} model. The performance for this case (referred to as `ours w/ ResNet') is reported in Table~\ref{tab:results}. The results using the relaxed criteria are shown in Figures~\ref{fig:knear} and~\ref{fig:editdist}. To analyze the effect of depth on the predictions, we also incorporated depth into the image tower. We add the depth image as another channel in the input layer of the image tower. For obtaining the depth images, we use the method of \cite{eigen14}, which estimates depth from a single image. We use their publicly available model, which is trained on a subset of the SUN RGB-D dataset. Using depth improves `ours w/ ResNet' and 'ours w/ AlexNet' by 2.9\% and 1.0\%, respectively (Table~\ref{tab:results}). It seems ResNet better leverages this additional source of information. We initialize the additional depth channel randomly (random samples from a Gaussian distribution with mean 0 and standard deviation 0.01). The results for these ablative cases using the relaxed criteria are also shown in Figure~\ref{fig:result}.

\begin{figure*}[t!]
\centering
  \includegraphics[width=29pc]{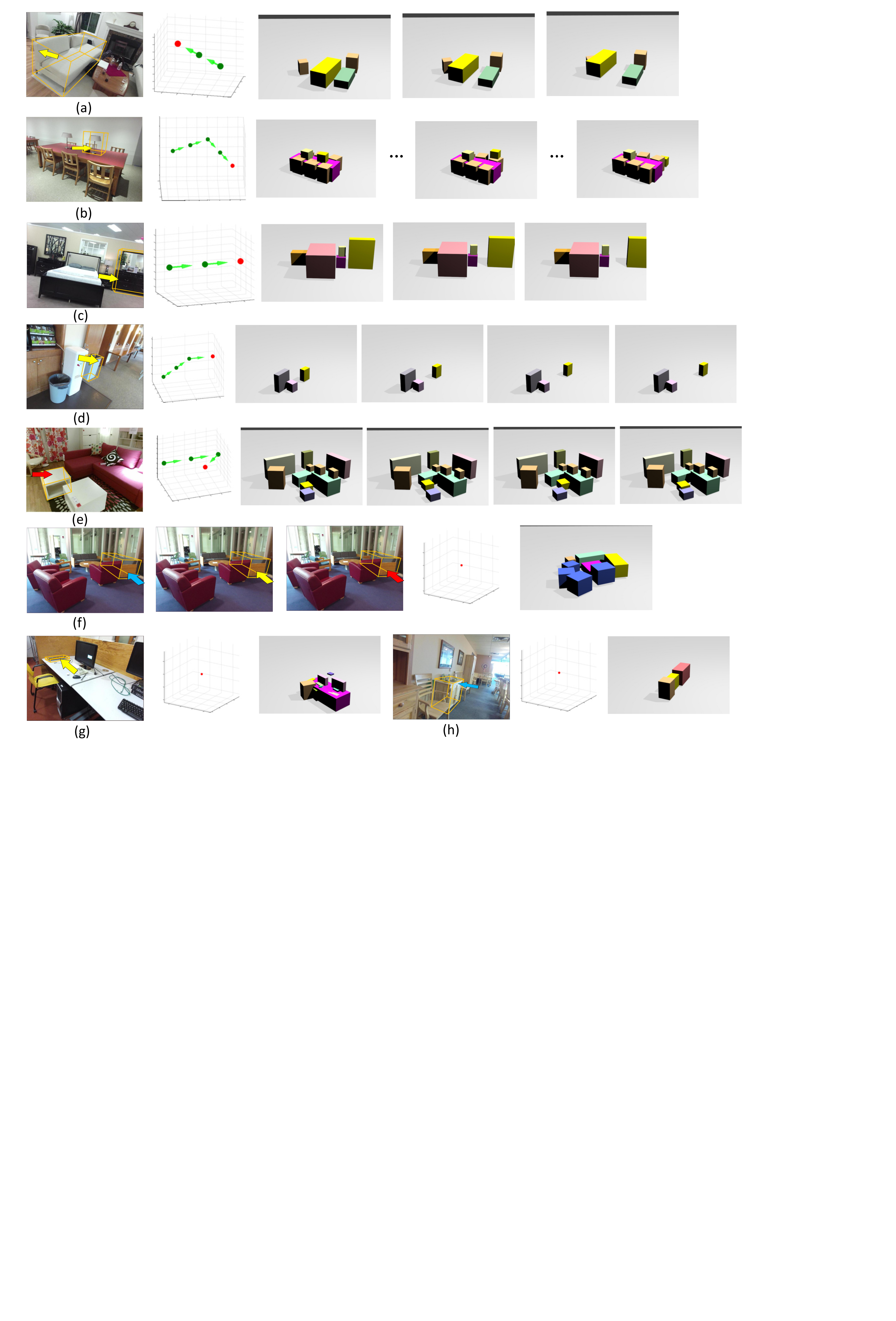}
\caption{\textbf{Qualitative results.} The left figure shows the force (color arrow) applied to the image. Different force magnitudes are shown with different colors, where blue, yellow, and red represent small, medium and large forces, respectively. The second image from the left shows the output of our method, which is a sequence of velocity vectors in 3D. The red point is the step that the velocity becomes zero. The resulted motion is visualized in the synthetic scenes. The object that moves is shown in yellow. Note that these synthetic scenes are for visualization purposes and they are not used during test. For clarity, we do not show walls.}
\label{fig:results}
\end{figure*}

\begin{figure*}[t]
\centering
  \includegraphics[width=28pc]{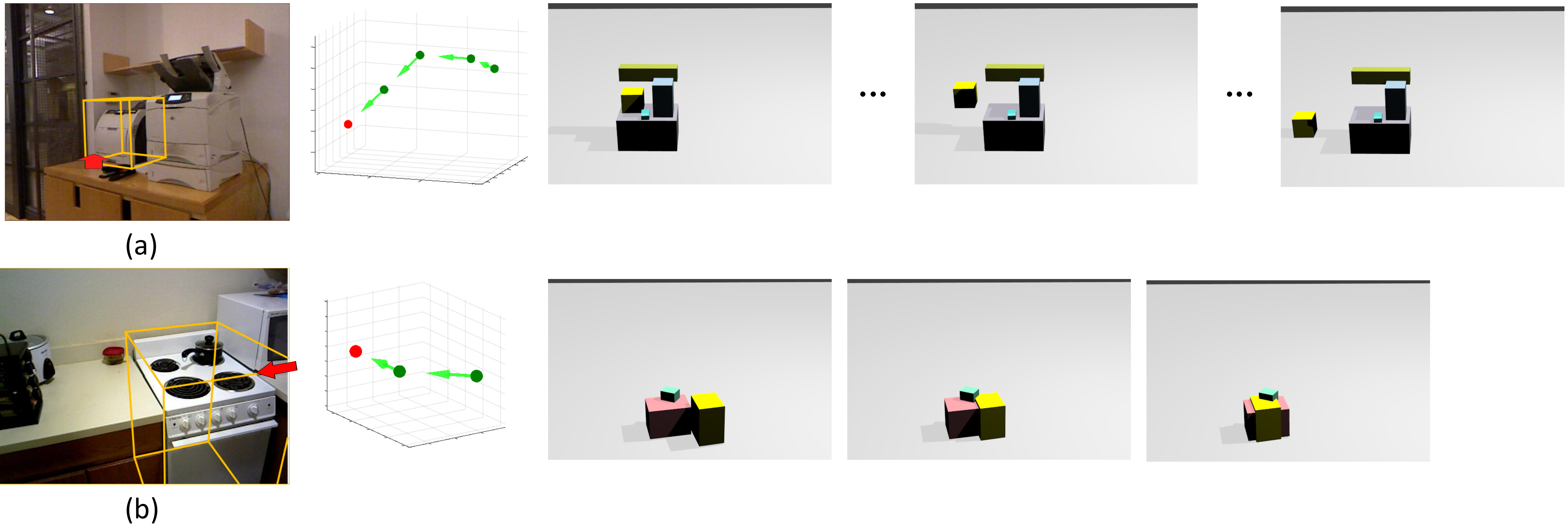}
\caption{\textbf{Failure cases.} For the details of the visualization, refer to the caption of Figure~\ref{fig:results}.}
\label{fig:failure}
\end{figure*}

Some qualitative results are shown in Figure~\ref{fig:results}. For example, Figures~\ref{fig:results}(a) and (c) show two cases that the object moves in the same direction as the force. Figure~\ref{fig:results}(b) shows an example of falling, where the lamp moves straight for two steps and then it drops. Figure~\ref{fig:results}(e) shows an example that the object bounces back as the result of applying a large force. Figure~\ref{fig:results}(f) shows an example that object does not move no matter how large the force is. It probably learns that pushing objects against a wall cannot cause a movement. There are two other examples in Figures~\ref{fig:results}(g) and (h), where the object does not move. We also show some failure cases in Figure~\ref{fig:failure}. In Figure~\ref{fig:failure}(a), the method ignores the wall behind the printer and infers a falling motion for the printer. In Figure~\ref{fig:failure}(b) the stove goes through the cabinet, which is not a correct prediction. Note that the synthetic scenes are just for visualization of the movements and they are not used during testing and inference. 

\noindent \textbf{Baseline methods.} The first baseline that we consider is a regression baseline, where we replace the RNN part of our network with a fully connected layer that maps $I$ (refer to Figure~\ref{fig:model}) to 18 numbers (we have at most 6 steps and at each step we want to predict a 3-dimensional vector). If the length of the training sequence is less than 6, we set their corresponding elements in the 18-dimensional vector to zero. We augment the network by a smooth L1 loss layer. As the result of regression, we obtain a vector of size 18, which corresponds to six 3-dimensional vectors. We assign them to different bins in the quantized direction space or the `stop' class (using the procedure described in Section~\ref{sec:subdataset}). The results are reported in Table~\ref{tab:results} and Figure~\ref{fig:result}. The result of the AlexNet-based regression method is 6.1\% lower than the result of `ours w/ AlexNet'. 

Another baseline that we tried is a nearest neighbor baseline. For each query object and force in the test set, we forward the corresponding RGB-M and the force image to the our full network (which is already trained using our data). We obtain the features $I$. Then, we find the query object and force in our training data that produces the most similar $I$. We use the sequence that is associated to the most similar training data as the predicted sequence. The features are high dimensional. Hence, to find the nearest neighbor we use multiple index hashing method of \cite{rastegari15}. The results of this AlexNet-based nearest neighbor is not competitive either (Table~\ref{tab:results} and Figure~\ref{fig:result}).

\begin{table}[t]
\setlength{\tabcolsep}{2pt}
\begin{center}
\begin{tabular}{|c|c|c|c|c|c|}
    \hline
	\begin{tabular}{@{}c@{}}ours w/ ResNet \\ + Depth\end{tabular} & ours w/ ResNet & \begin{tabular}{@{}c@{}}ours w/ AlexNet \\ + Depth\end{tabular} & ours w/ AlexNet & \begin{tabular}{@{}c@{}} Regression \\ AlexNet \end{tabular}& \begin{tabular}{@{}c@{}}Nearest Neigh. \\ AlexNet\end{tabular}\\
	\hline
    \textbf{19.8} & 16.9 & 17.5 & 16.5 & 10.4 & 8.5 \\
    \hline
\end{tabular}
\end{center}
\caption{Ablative analysis of our method and comparison with baseline approaches. The evaluation metric is the percentage of sequences that we predict correctly.}
\label{tab:results}
\end{table}

\begin{table}[t]
\setlength{\tabcolsep}{3pt}
\begin{center}
\begin{tabular}{|c|c|c|c|c|c|c|c|c|c|c||c|}
    \hline
	chair & table & desk & pillow & sofa chair & sofa & bed & box & garbage bin & shelf & \textbf{Avg.} & \textbf{All}\\ 
	\hline
17.7 & 17.0 & 15.4 & 15.5 & 15.9 & 17.2 & 15.9 & 14.6 & 16.1 & 15.9 & 16.12 & 16.53 \\
    \hline
\end{tabular}
\end{center}
\caption{Generalization of the method to the classes that were not seen during training. Each column shows the results for the case that we remove the sequences corresponding to that category from the training set. The rightmost column (`All') shows the base case, where all training examples are seen.}
\label{tab:zeroshot}
\end{table}

\subsection{Unseen categories}
To evaluate how well our method generalizes to object categories that are not seen during training, we remove the training sequences that correspond to an object category and evaluate the method on the entire test set. For this experiment, we consider the ten most frequent object categories in our dataset. 

We re-train the network each time we remove the sequences corresponding to an object category from our training set. The result of this experiment is shown in Table~\ref{tab:zeroshot}. We report the results using the strict evaluation criteria. We use the method that we refer to as `ours w/ AlexNet' for this experiment since its training time is faster than our other approaches. The results show that the average performance does not drop significantly compared to the case that we use the entire training set. This means that our method generalizes well to the categories that it has not seen during training. 

\section{Conclusion}
Visual reasoning is a key component of any intelligent agent that is supposed to operate in the visual world. An important component in visual reasoning is the ability to predict the expected outcome of an action. This capability enables planing, reasoning about actions, and eventually successfully executing tasks. In this paper, we take one step toward this crucial component and study the problem of prediction the effect of an action (represented as a force vector) when applied to an object in an image. Our experimental evaluations show that our model can, in fact, predict long-term sequential movements of objects when a force is applied to them. 

Our solution is mainly concerned with predicting translation vectors and does not take into account rotation of objects around their centers. Extending our model to also predict the rotations would be straightforward. Also our current model assumes uniform weights for all objects, resulting in a calibration issue for the magnitude of the force necessary to move an object. Large scale estimation of weights of objects from visual data is an interesting future direction. Considering the current success of implicit approaches in recognition, we also adopt an implicit approach to our problem. We found explicit estimation of the components involved in the prediction of the physical progression of objects to be challenging. In fact, our initial experiments show that reliable estimation of many of the geometrical and physical properties of objects is still beyond the state of the art. We used an implicit model to directly learn the end goal and estimate the necessary components implicitly. Exploring the re-estimation of physical and geometrical properties from visual data and the predicted movements is another interesting research direction enabled by this work.  

\section*{Appendix}
\begin{enumerate}
\item We applied the forces to the instances of the following 50 categories in the SUN RGB-D dataset: chair, table, desk, pillow, sofa chair, sofa, bed, box, garbage bin, shelf, lamp, cabinet, computer, monitor, drawer, endtable, night stand, bookshelf, dresser, keyboard, picture, coffee table, whiteboard, toilet, paper, tv, ottoman, fridge, printer, bench, bottle, laptop,    book, bag, stove, papers, cup, backpack, dresser mirror, tray, bowl, mouse, telephone, mug, cart, speaker, scanner, suitcase, flower vase, and coffee maker.

\item The following video shows some of our results and also a number of failure cases: \burl{https://s3-us-west-2.amazonaws.com/ai2-vision-datasets/prediction_videos/force_results.mp4}
\end{enumerate}

\bibliographystyle{splncs03}
\bibliography{egbib}

\end{document}